\def\year{2019}\relax
\documentclass[letterpaper]{article} 
\usepackage{include/aaai19}  
\usepackage{times}  
\usepackage{helvet}  
\usepackage{courier}  
\PassOptionsToPackage{hyphens}{url}\usepackage{hyperref}  
\usepackage{graphicx}  
\usepackage{color}
\usepackage[html]{xcolor} 
\usepackage[utf8]{inputenc}
\usepackage{booktabs}
\usepackage{natbib}
\usepackage{tikz}
\usetikzlibrary{arrows, positioning, 3d, calc, fit}
\usepackage{tabularx}

\usetikzlibrary{external}

\usepackage{amsmath,amssymb}
\DeclareMathOperator{\E}{\mathbb{E}}

\renewcommand{\Re}{\operatorname{Re}}
\renewcommand{\Im}{\operatorname{Im}}

\usepackage{nameref}
\newcommand{\secref}[1]{\textit{\nameref{#1}}}

\usepackage{subcaption}
\usepackage{cleveref}  

\tikzstyle{perspective3d}=[z={(0.5cm,-0.5cm)}, x={(1cm,0cm)}, y={(0cm,1cm)}]
\tikzstyle{conn}=[-stealth, rounded corners]
\tikzstyle{revconn}=[stealth-, rounded corners]
\tikzstyle{shared}=[stealth-stealth, rounded corners, very thick]
\tikzstyle{fm}=[inner sep=0]
\tikzstyle{net}=[inner sep=.5em, draw=gridcolor, rounded corners]
\tikzstyle{grid}=[inner sep=0]

\newcommand{\mygridd}[4]{%
        \def\scale{#1}
        \def\depth{#2}
        \def\step{#3}
        \def\fillcolor{#4}

        \begin{tikzpicture}[perspective3d, scale=\scale]
            
            \draw[canvas is xy plane at z=1, fill=\fillcolor] (0,0) -- (0,1) -- (\depth,1) -- (\depth,0) -- (0,0);
            \draw[canvas is xz plane at y=1, fill=\fillcolor] (0,0) -- (0,1) -- (\depth,1) -- (\depth,0) -- (0,0);
            \draw[canvas is yz plane at x=0, fill=\fillcolor] (0,0) -- (0,1) -- (1,1) -- (1,0) -- (0,0);
            \draw[canvas is yz plane at x=0, step=\step, draw=black] (0,0)  grid (1,1);
            
        \end{tikzpicture}    
        }%

\newcommand{\concatsone}{
\begin{tikzpicture}
    \node[inner sep=0] at (0,0){\mygridd{1}{.125}{.25}{activationcolor!70}};
    \node[inner sep=0] at (-.1,0){\mygridd{1}{.125}{.25}{activationcolor}};
\end{tikzpicture}
}

\newcommand{\concatstwo}{
\begin{tikzpicture}
    \node[inner sep=0] at (0,0){\mygridd{1}{.125}{.33}{activationcolor!70}};
    \node[inner sep=0] at (-.1,0){\mygridd{1}{.125}{.33}{activationcolor}};
\end{tikzpicture}
}

\newcommand{\concatcontext}{
\begin{tikzpicture}
    \node[inner sep=0] at (0,0){\mygridd{1}{.5}{.25}{activationcolor}};
    \node[inner sep=0] at (-.1,0){\mygridd{1}{.125}{.25}{contextonecolor}};
    \node[inner sep=0] at (-.2,0){\mygridd{1}{.125}{.25}{contexttwocolor}};
    \node[inner sep=0] at (-.3,0){\mygridd{1}{.125}{.25}{contextthreecolor}};
    \node[inner sep=0] at (-.4,0){\mygridd{1}{.125}{.25}{contextfourcolor}};
\end{tikzpicture}
}

\newcommand{\contextfeaturemaps}{
\begin{tikzpicture}[xshift=1]
    \node[inner sep=0] at (-.1,0){\mygridd{1}{.125}{.25}{contextonecolor}};
    \node[inner sep=0] at (-.2,0){\mygridd{1}{.125}{.25}{contexttwocolor}};
    \node[inner sep=0] at (-.3,0){\mygridd{1}{.125}{.25}{contextthreecolor}};
    \node[inner sep=0] at (-.4,0){\mygridd{1}{.125}{.25}{contextfourcolor}};
\end{tikzpicture}
}

\newcommand{\figpspmodule}{
    \begin{tikzpicture}[node distance=.5em, xscale=1.2, yscale=-1.25]
    
    \coordinate (legend) at (0,-.5);

    \node[label=below:features](input) at (-1.8,1.5) {\mygridd{1}{.5}{.25}{activationcolor}};
    \node[label=below:context, inner sep=0](output) at (3.5,1.5) {\contextfeaturemaps};

    \coordinate[right=1.2em of input] (pool);
    \coordinate[left=1em of output] (upsample);
    
    \node[grid](grid1) at (0.2,0){\mygridd{.3}{.5}{.5}{activationcolor!40}};
    \node[grid](grid2) at (0.2,.8){\mygridd{.5}{.5}{.25}{activationcolor!60}};
    \node[grid](grid3) at (0.2,1.8){\mygridd{.75}{.5}{.25}{activationcolor!80}};
    \node[grid](grid16) at (0.2,3){\mygridd{1}{.5}{.25}{activationcolor}};
    
    \coordinate[below right= 1em and 2em of grid16](convcoord);
    
    \node[grid](gridd1) at (1.8,0){\mygridd{.3}{.2}{.5}{contextonecolor}};
    \node[grid](gridd2) at (1.8,.8){\mygridd{.5}{.2}{.25}{contexttwocolor}};
    \node[grid](gridd3) at (1.8,1.8){\mygridd{.75}{.2}{.25}{contextthreecolor}};
    \node[grid](gridd16) at (1.8,3){\mygridd{1}{.2}{.25}{contextfourcolor}};

    \coordinate (center) at ($ (gridd1)!0.5!(grid1) $);
    
    \draw[conn] (input) -- (pool) |- (grid1);
    \draw[conn] (input) -- (pool) |- (grid2);
    \draw[conn] (input) -- (pool) |- (grid3);
    \draw[conn] (input) -- (pool) |- (grid16);
    
    \draw[conn] (grid1) -- (gridd1);
    \draw[conn] (grid2) -- (gridd2);
    \draw[conn] (grid3) -- (gridd3);
    \draw[conn] (grid16) -- (gridd16);
    
    \draw[revconn] (output) -- (upsample) |- (gridd1);
    \draw[revconn] (output) -- (upsample) |- (gridd2);
    \draw[revconn] (output) -- (upsample) |- (gridd3);
    \draw[revconn] (output) -- (upsample) |- (gridd16);

    \node[anchor=mid](legpool) at (legend -| pool) {pooling};
    \node[anchor=mid](legupsampl) at (legend -| upsample) {upsampling};
    \node[anchor=mid](legconv) at (legend -| center) {convolution};
    
    \end{tikzpicture}
}

\newcommand{\figFusionNetwork}{
\begin{tikzpicture}[node distance=1em, yscale=1.35]

    \def\labelimage{images/tile/networkoutputbuildings}
    
    \node[label={[anchor=south, rotate=90,label distance=0em]left:\small pre}] at (0,3.03) (s1pre) {\includegraphics[width=1.2cm]{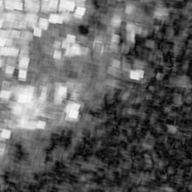}};
    \node[label={[anchor=south, rotate=90,label distance=0em]left:\small post}] at (0,3.97) (s1post) {\includegraphics[width=1.2cm]{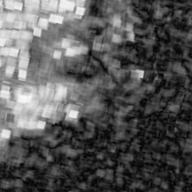}};
    \node[anchor=south, rotate=90] at (-1.3,3.5) {Sentinel-1};

    \coordinate (center) at ($ (s1pre.east)!0.5!(s1post.east) $);
    \node[right=of center, inner sep=0, label={[anchor=center, rotate=90,label distance=0em]left:\scriptsize 192px}] (minus) {\concatsone};

    \node[net,right=of minus] (resnet) {CNN encoder};
    \node[right=of resnet, inner sep=0, label={[anchor=center, label distance=.4em]below:\scriptsize 96px}] (h0) {\mygridd{1}{.5}{.25}{activationcolor}};
    \node[net, right=of h0, align=left] (pspmodule) {Context Module};
    \node[right=of pspmodule, inner sep=0, label={[anchor=center, label distance=.4em]below:\scriptsize 96px}] (aggrout) {\concatcontext};
    \node[right=of aggrout] (s1label) {\includegraphics[width=1.2cm]{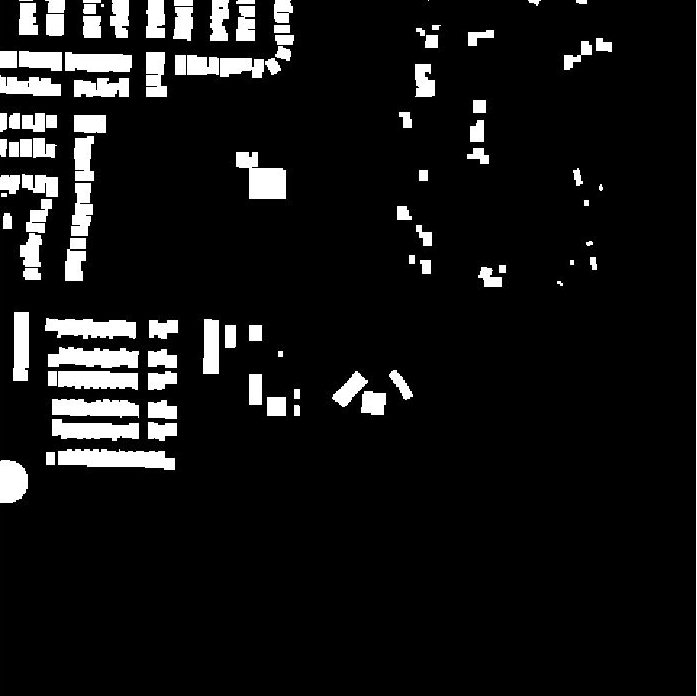}};

    \draw[conn] (s1pre.east)++(0,-.8em) -| (minus);
    \draw[conn] (s1post.east)++(0,.8em) -| (minus);
    \draw[conn] (minus) -- (resnet);
    \draw[conn] (resnet) -- (h0);
    \draw[conn] (h0) -- (pspmodule);
    \draw[conn] (pspmodule) -- (aggrout);
    \draw[conn] (aggrout) -- (s1label);
    \draw[conn] (h0) |- ++(1,2em) -| (aggrout);

    \node[label={[label distance=0em, anchor=south, rotate=90]left:\small pre}] at (0,1.03) (s2pre) {\includegraphics[width=1.2cm]{images/tile/s2pre}};
    \node[label={[anchor=south, rotate=90,label distance=0em]left:\small post}] at (0,1.97) (s2post) {\includegraphics[width=1.2cm]{images/tile/s2post}};
    \node[anchor=south, rotate=90] at (-1.3,1.5) {Sentinel-2};

    \coordinate (center) at ($ (s2pre.east)!0.5!(s2post.east) $);
    \node[right=of center, inner sep=0, label={[anchor=center, rotate=90,label distance=0em]left:\scriptsize 96px}] (minus) {\concatstwo};
    \node[net,right=of minus] (resnet) {CNN encoder};
    \node[right=of resnet, inner sep=0, label={[anchor=center, label distance=.4em]below:\scriptsize 96px}] (h0) {\mygridd{1}{.5}{.25}{activationcolor}};
    \node[net, right=of h0, align=left] (pspmodule) {Context Module};
    \node[right=of pspmodule, inner sep=0, label={[anchor=center, label distance=.4em]below:\scriptsize 96px}] (aggrout) {\concatcontext};
    \node[right=of aggrout] (s2label) {\includegraphics[width=1.2cm]{\labelimage}};

    \draw[conn] (s2pre.east)++(0,-.8em) -| (minus);
    \draw[conn] (s2post.east)++(0,.8em) -| (minus);
    \draw[conn] (minus) -- (resnet);
    \draw[conn] (resnet) -- (h0);
    \draw[conn] (h0) -- (pspmodule);
    \draw[conn] (pspmodule) -- (aggrout);
    \draw[conn] (aggrout) -- (s2label);
    \draw[conn] (h0) |- ++(1,2em) -| (aggrout);

    \node[label={[anchor=south, rotate=90,label distance=0em]left:\small post}] (input) at (0,-.15) {\includegraphics[width=1.4cm]{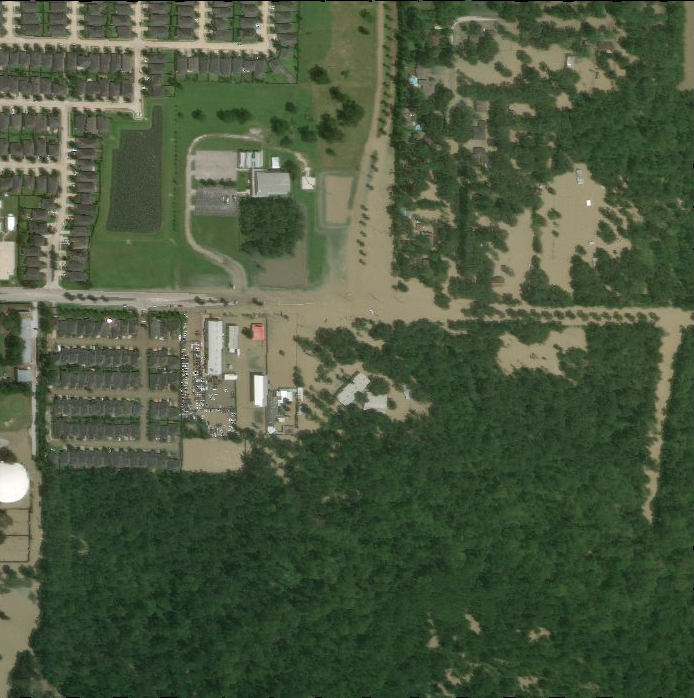}};
    \node[right=of input, inner sep=0, label={[anchor=center, label distance=.4em]below:\scriptsize 1560px}] (hm1) {\mygridd{1}{.125}{.125}{activationcolor}};
    \node[net,right=of hm1] (resnet) {CNN encoder};
    \node[right=of resnet, inner sep=0, label={[anchor=center, label distance=.4em]below:\scriptsize 96px}] (h0) {\mygridd{1}{.5}{.25}{activationcolor}};
    \node[net,right=of h0, align=left] (pspnet) {Context Module};
    \node[right=of pspnet, inner sep=0, label={[anchor=center, label distance=.4em]below:\scriptsize 96px}] (aggrout) {\concatcontext};
    \node[right=of aggrout, align=left] (pspout) {\includegraphics[width=1.2cm]{\labelimage}};
    
    \draw[conn] (input.east) -- (hm1);
    \draw[conn] (hm1) -- (resnet);
    \draw[conn] (resnet) -- (h0);
    \draw[conn] (h0) -- (pspnet);
    \draw[conn] (pspnet) -- (aggrout);
    \draw[conn] (aggrout) -- (pspout);
    \draw[conn] (h0) |- ++(1,2em) -| (aggrout);

    \node[anchor=south, rotate=90] at (-1.3,-.15) {VHR};

    \node[right=2em of s2label, label=below:prediction](out){\includegraphics[width=1.5cm]{\labelimage}};
    
    \draw[conn] (s2label) -- (out);
    \draw[conn] (s1label) -| ($ (s1label.east)!0.5!(out.west) $) |- (out.west);
    \draw[conn] (pspout) -| ($ (pspout.east)!0.5!(out.west) $) |- (out.west);
    
\end{tikzpicture}
}

\definecolor{gridcolor}{HTML}{284251}

\definecolor{focusone}{HTML}{4575b4} 
\definecolor{focustwo}{HTML}{be3e2b} 

\colorlet{activationcolor}{focusone} 
\colorlet{contextonecolor}{focustwo!40}
\colorlet{contexttwocolor}{focustwo!60}
\colorlet{contextthreecolor}{focustwo!80}
\colorlet{contextfourcolor}{focustwo}

\begin{document}

\newcommand{\plaincopy}[1]{{\color{blue}#1}}

\makeatletter
\def\blfootnote{\xdef\@thefnmark{}\@footnotetext}
\makeatother

\title{Multi$^\textbf{3}$Net: Segmenting Flooded Buildings via Fusion of Multiresolution, Multisensor, and Multitemporal Satellite Imagery}

 \author{
  	\hspace{35pt}Tim G. J. Rudner$^{\dagger}$ \\
 	\hspace{35pt}University of Oxford \\
 	\hspace{35pt}tim.rudner@cs.ox.ac.uk \\
 	\And
 	\hspace{35pt}Marc Ru\ss{}wurm$^{\dagger}$ \\
 	\hspace{35pt}TU Munich \\
 	\hspace{35pt}marc.russwurm@tum.de \\
 	\And
 	Jakub Fil$^{\dagger}$ \\
 	University of Kent \\
 	jf330@kent.ac.uk \\
 	\And
 	\hspace{-35pt}Ramona Pelich$^{\dagger}$ \\
 	\hspace{-35pt}LIST Luxembourg \\
 	\hspace{-35pt}ramona.pelich@list.lu \\
 	\And
 	\hspace{-35pt}Benjamin Bischke$^{\dagger}$ \\
 	\hspace{-35pt}DFKI \& TU Kaiserslautern \\
 	\hspace{-35pt}benjamin.bischke@dfki.de\\
 	\AND
 	Veronika Kopačková \\
 	Czech Geological Survey \\
 	veronika.kopackova@seznam.cz \\
 	\And
 	Piotr Biliński \\
 	University of Oxford \& University of Warsaw \\
 	piotrb@robots.ox.ac.uk
}

\maketitle

\begin{abstract}

We propose a novel approach for rapid segmentation of flooded buildings by fusing multiresolution, multisensor, and multitemporal satellite imagery in a convolutional neural network.
Our model significantly expedites the generation of satellite imagery-based flood maps, crucial for first responders and local authorities in the early stages of flood events.
By incorporating multitemporal satellite imagery, our model allows for rapid and accurate post-disaster damage assessment and can be used by governments to better coordinate medium- and long-term financial assistance programs for affected areas.
The network consists of multiple streams of encoder-decoder architectures that extract spatiotemporal information from medium-resolution images and spatial information from high-resolution images before fusing the resulting representations into a single medium-resolution segmentation map of flooded buildings.
We compare our model to state-of-the-art methods for building footprint segmentation as well as to alternative fusion approaches for the segmentation of flooded buildings and find that our model performs best on both tasks.
We also demonstrate that our model produces highly accurate segmentation maps of flooded buildings using only publicly available medium-resolution data instead of significantly more detailed but sparsely available very high-resolution data.
We release the first open-source dataset of fully preprocessed and labeled multiresolution, multispectral, and multitemporal satellite images of disaster sites along with our source code.

\end{abstract}


\section{Introduction}
\blfootnote{$^\dagger$Equal contribution.}

In 2017, Houston, Texas, the fourth largest city in the United States, was hit by tropical storm Harvey, the worst storm to pass through the city in over 50 years.
Harvey flooded large parts of the city, inundating over 154,170 homes and leading to more than 80 deaths. 
According to the US National Hurricane Center, the storm caused over 125 billion USD in damage, making it the second-costliest storm ever recorded in the United States.
Floods can cause loss of life and substantial property damage. Moreover, the economic ramifications of flood damage disproportionately impact the most vulnerable members of society.

When a region is hit by heavy rainfall or a hurricane, authorized representatives of national civil protection, rescue, and security organizations can activate the International Charter `Space and Major Disasters'. Once the Charter has been activated, various corporate, national, and international space agencies task their satellites to acquire imagery of the affected region.
As soon as images are obtained, satellite imagery specialists visually or semi-automatically interpret them to create flood maps to be delivered to disaster relief organizations.
Due to the semi-automated nature of the map generation process, delivery of flood maps can take several hours after the imagery was provided.

We propose Multi$^3$Net, a novel approach for rapid and accurate flood damage segmentation by fusing multiresolution and multisensor satellite imagery in a convolutional neural network (CNN). 
The network consists of multiple deep encoder-decoder streams, each of which produces an output map based on data from a single sensor. If data from multiple sensors is available, the streams are combined into a joint prediction map.
We demonstrate the usefulness of our model for segmentation of flooded buildings as well as for conventional building footprint segmentation.

Our method aims to reduce the amount of time needed to generate satellite imagery-based flood maps by fusing images from multiple satellite sensors.
Segmentation maps can be produced as soon as at least a single satellite image acquisition has been successful and subsequently be improved upon once additional imagery becomes available.
This way, the amount of time needed to generate satellite imagery-based flood maps can be reduced significantly, helping first responders and local authorities make swift and well-informed decisions when responding to flood events. Additionally, by incorporating multitemporal satellite imagery, our method allows for a speedy and accurate post-disaster damage assessment, helping governments better coordinate medium- and long-term financial assistance programs for affected areas.

The main contributions of this paper are (1) the development of a new fusion method for multiresolution, multisensor, and multitemporal satellite imagery and (2) the creation and release of a dataset containing labeled multisensor and multitemporal satellite images of multiple disaster sites.\footnote{\scriptsize\url{https://github.com/FrontierDevelopmentLab/multi3net}.}

\begin{figure*}
\begin{subfigure}{.196\textwidth}\includegraphics[width=\textwidth]{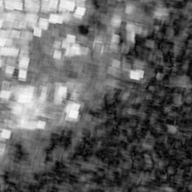}\caption{Sentinel-1 (192px) \\ coherence pre-event}\label{fig:tile:s1pre}\end{subfigure}
\begin{subfigure}{.196\textwidth}\includegraphics[width=\textwidth]{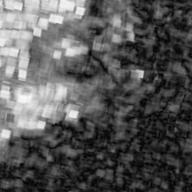}\caption{Sentinel-1 (192px)\\ coherence post-event}\label{fig:tile:s1post}\end{subfigure}
\begin{subfigure}{.196\textwidth}\includegraphics[width=\textwidth]{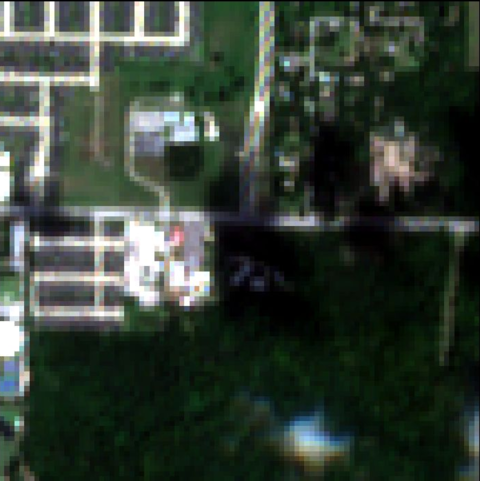}\caption{Sentinel-2 (96px)\\ pre-event}\label{fig:tile:s2pre}\end{subfigure}
\begin{subfigure}{.196\textwidth}\includegraphics[width=\textwidth]{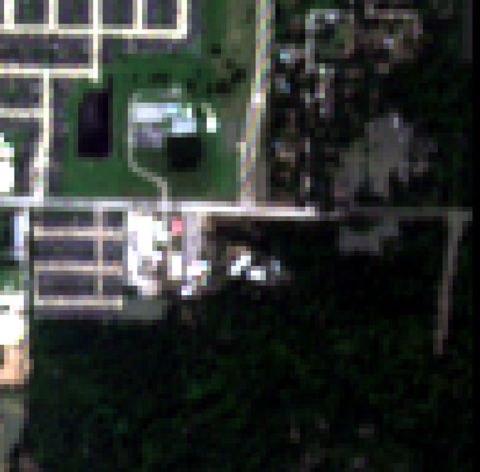}\caption{Sentinel-2 (96px) \\ post-event}\label{fig:tile:s2post}\end{subfigure}
\begin{subfigure}{.196\textwidth}\includegraphics[width=.97\textwidth]{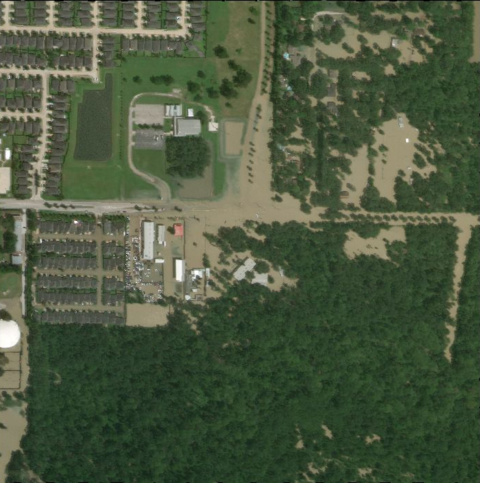}\caption{Very high-res. (1560px) \\ post-event}\label{fig:tile:vhr}\end{subfigure}
\caption{
One image tile of 960m$\times$960m is used as network input.
Figures (a) and (b) illustrate Sentinel-1 coherence images before and after the flood event, whereas Figures (c) and (d) show RGB representations of multispectral Sentinel-2 optical images.
Figure (e) shows the high level of spatial details in a very high-resolution image.
While the medium-resolution (Sentinel-1 and Sentinel-2) images contain temporal information, the very high-resolution image encodes more spatial detail.
}
\label{fig:tile}
\end{figure*}

\section{Background: Earth Observation}
\label{sec:satellitedata}

There is an increasing number of satellites monitoring the Earth's surface, each designed to capture distinct surface properties and to be used for a specific set of applications. 
Satellites with optical sensors acquire images in the visible and short-wavelength parts of the electromagnetic spectrum that contain information about chemical properties of the captured scene.
Satellites with radar sensors, in contrast, use longer wavelengths than those with optical sensors, allowing them to capture physical properties of the Earth’s surface \citep{Soergel2010}.
Radar images are widely used in the fields of \textit{Earth observation} and \textit{remote sensing}, since radar image acquisitions are unaffected by cloud coverage or lack of light \citep{ulaby2014}.
Examples of medium- and very high-resolution optical and medium-resolution radar images are shown in \Cref{fig:tile}.

Remote sensing-aided disaster response typically uses very high-resolution (VHR) optical and radar imagery.
Very high-resolution optical imagery with a ground resolution of less than 1m is visually-interpretable and can be used to manually or automatically extract locations of obstacles or damaged objects.
Satellite acquisitions of very high-resolution imagery need to be scheduled and become available only after a disaster event.
In contrast, satellites with medium-resolution sensors of 10m--30m ground resolution monitor the Earth's surface with weekly image acquisitions for any location globally.
Radar sensors are often used to map floods in sparsely built-up areas since smooth water surfaces reflect electromagnetic waves away from the sensor, whereas buildings reflect them back.
As a result, conventional remote sensing flood mapping models perform poorly on images of urban or suburban areas.


\section{Related Work}

Recent advances in computer vision and the rapid increase of commercially and publicly available medium- and high-resolution satellite imagery have given rise to a new area of research at the interface of machine learning and remote sensing, as summarized by \citet{ZhuTuia2018RSM} and \citet{Zhang2016DeepLF}. 

One popular task in this domain is the segmentation of building footprints from satellite imagery, which has led to competitions such as the DeepGlobe \citep{demir2018DeepGlobe} and SpaceNet challenges \citep{van2018spacenet}.
Encoder-decoder networks like U-Net and SegNet are consistently among the best-performing models at such competitions and considered state-of-the-art for satellite imagery-based image segmentation \citep{Bischke2017MultiTaskLF, lexieyang2018}.
U-Net-based approaches that replace the original VGG architecture \citep{simonyan2014VGG} with, for example, ResNet encoders \citep{he2016resnet} performed best at the 2018 DeepGlobe challenge \citep{hamaguchi2018building}.
Recently developed computer vision models, such as DeepLab-v3 \citep{chen2017rethinking}, PSPNet \citep{zhao2017pyramid}, or DDSC \citep{Bilinski_2018_CVPR}, however, use improved encoder architectures with a higher receptive field and additional context modules.

Segmentation of damaged buildings is similar to segmentation of building footprints.
However, the former can be more challenging than the latter due to the existence of additional, confounding features, such as damaged non-building structures, in the image scene.
Adding a temporal dimension by using pre- and post-disaster imagery can help improve the accuracy of damaged building segmentation. For instance, \citet{Cooner2016DetectionOU} insert pairs of pre- and post-disaster images into a feedforward neural network and a random forest model, allowing them to identify buildings damaged by the 2010 Haiti earthquake. 
\Citet{Scarsi2014AnAF}, in contrast, apply an unsupervised method based on a Gaussian finite mixture model to pairs of very high-resolution WorldView-2 images and use it to assess the level of damage after the 2013 Colorado flood through change segmentation modeling.
If pre- and post-disaster image pairs of the same type are unavailable, it is possible to combine different image types, such as optical and radar imagery. \Citet{Brunner2010EarthquakeDA}, for example, use a Bayesian inference method to identify collapsed buildings after an earthquake from pre-event very high-resolution optical and post-event very high-resolution radar imagery.

There are other methods, however, which only rely on post-disaster images and data augmentation.
\Citet{Bai2018AFO} use data augmentation to generate a training dataset for deep neural networks, enabling rapid segmentation of building footprints in satellite images acquired after the 2011 Tohoku earthquake and tsunami in Japan.


\section{Method}

\begin{figure}[t]
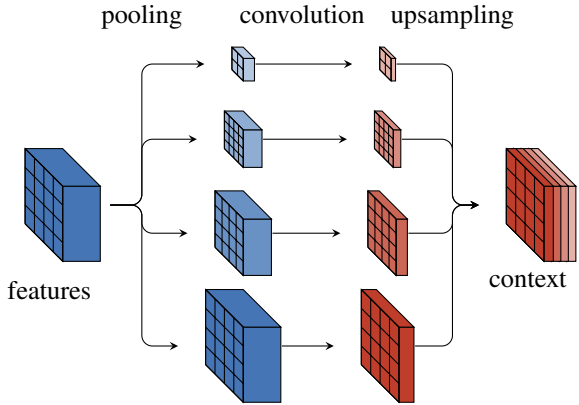

    \begin{centering}
    	\tikzsetnextfilename{pspmodule}
		\figpspmodule
    \end{centering}
    \caption{Multi$^3$Net's context aggregation module extracts and combines image features at different image resolutions, similarly to \citet{zhao2017pyramid}.}  
    \label{fig:pspmodule}
\end{figure}

In this section, we introduce Multi$^3$Net, an approach to segmenting flooded buildings using multiple types of satellite imagery in a multi-stream convolutional neural network.
We first describe the architecture of our segmentation network for processing images from a single satellite sensor. 
Building on this approach, we propose an extension to the network, which allows us to effectively combine information from different types of satellite imagery, including multiple sensors and resolutions across time.


\subsection{Segmentation Network Architecture}

Multi$^3$Net uses an encoder-decoder architecture. 
In particular, we use a modified version of ResNet \citep{he2016resnet} with dilated convolutions as feature extractors \citep{yu2017dilated} that allows us to effectively downsample the input image along the spatial dimensions by a factor of only $\times8$ instead of $\times32$. 
Motivated by the recent success of multi-scale features \citep{zhao2017pyramid, chen2017rethinking}, we enrich the feature maps with an additional context aggregation module as depicted in \Cref{fig:pspmodule}. 
This addition to the network allows us to incorporate contextual image information into the encoded image representation. 
The decoder component of the network uses three blocks of bilinear upsampling functions with a factor of $\times2$, followed by a 3$\times$3 convolution, and a PReLU activation function to learn a mapping from latent space to label space. 
The network is trained end-to-end using backpropagation.

\subsection{Multi$^\textbf{3}$Net Image Fusion}

\begin{figure*}
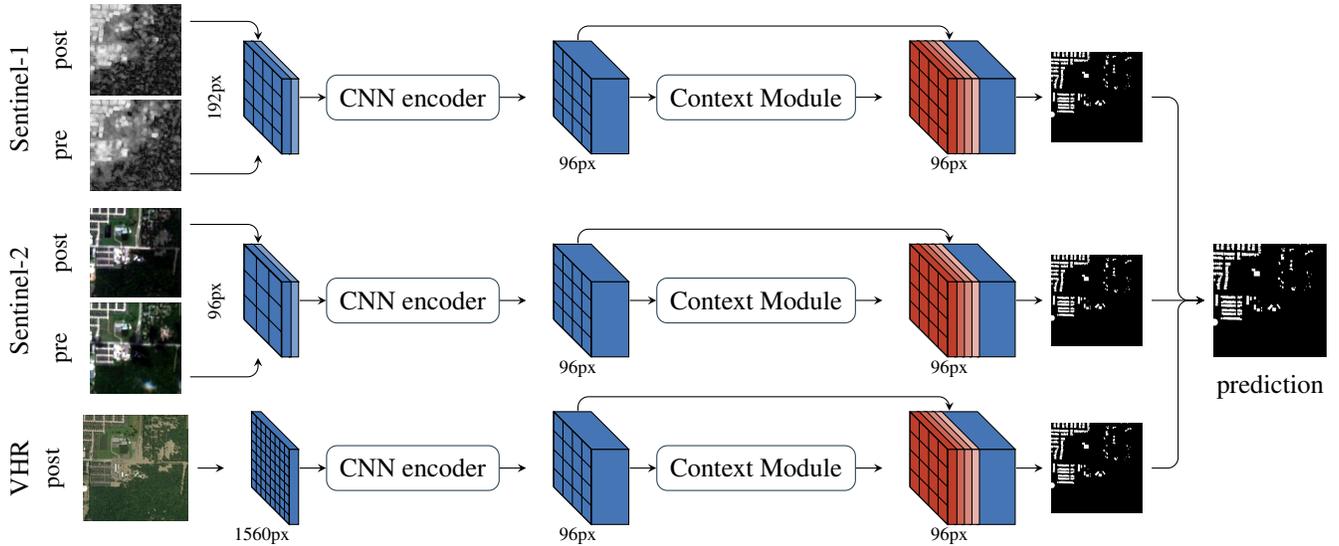

\begin{center}
    \begin{subfigure}{\textwidth}
    	\tikzsetnextfilename{fusionnetwork}
		\figFusionNetwork
    \end{subfigure}
    \caption{Overview of Multi$^3$Net's multi-stream architecture. Each satellite image is processed by a separate stream that extracts feature maps using a CNN-encoder and then augments them with contextual features. Features are mapped to the same spatial resolution, and the final prediction is obtained by fusing the predictions of individual streams using additional convolutions.}
	\label{fig:pspfusion}
\end{center}
\end{figure*}

Multi$^3$Net fuses images obtained at multiple points in time from multiple sensors with different resolutions to capture different properties of the Earth's surface across time. In this section, we address each fusion type separately.


\subsubsection{Multisensor Fusion}

Images obtained from different sensors can be fused using a variety of approaches. We consider \textit{early} as well as \textit{late-fusion}.
In the early-fusion approach, we upsample each satellite image, concatenate them into a single input tensor, and then process the information within a single network.
In the late-fusion approach, each image type is fed into a dedicated information processing stream as shown in the segmentation network architecture depicted in \Cref{fig:pspfusion}. 
We first extract features separately from each satellite image and then combine the class predictions from each individual stream by first concatenating them and then applying additional convolutions. 
We compared the performance of several network architectures, fusing the feature maps in the encoder (as was done in FuseNet \citep{hazirbas2016fusenet}) and using different late-fusion approaches, such as sum fusion or element-wise multiplication, and found that a late-fusion approach, in which the output of each stream is fused using additional convolutional layers, achieved the best performance.
This finding is consistent with related work on computer vision focused on the fusion of RGB optical images and depth sensors \citep{couprie2013indoor}. 
In this setup, the segmentation maps from the different streams are fused by concatenating the segmentation map tensors and applying two additional layers of 3$\times$3 convolutions with PReLU activations and a 1$\times$1 convolution. This way, the dimensions along the channels can be reduced until they are equal to the number of class labels.


\subsubsection{Multiresolution Fusion} In order to best incorporate the satellite images' different spatial resolutions, we follow two different approaches. When only Sentinel-1 and Sentinel-2 images are available, we transform the feature maps into a common resolution of $96\text{px} \times 96$px at a 10m ground resolution by removing one upsampling layer in the Sentinel-2 encoder network. Whenever very high-resolution optical imagery is available as well, we also remove the upsampling layer in the very high-resolution subnetwork to match the feature maps of the two Sentinel imagery streams.


\subsubsection{Multitemporal Fusion}
To quantify changes in the scene shown in a satellite images over time, we use pre- and post-disaster satellite images. We achieved the best results by concatenating both images into a single input tensor and processing them in the early-fusion network described above. More complex approaches, such as using two-stream networks with shared encoder weights similar to Siamese networks \citep{melekhov2016siamese} or subtracting the activations of the feature maps, did not improve model performance.


\subsection{Network Training}

We initialize the encoder with the weights of a ResNet34 model \citep{he2016resnet} pre-trained on ImageNet \citep{imagenet_cvpr09}. 
When there are more than three input channels in the first convolution (due to the 10 spectral bands of the Sentinel-2 satellite images), we initialize additional channels with the average over the first convolutional filters of the RGB channels. 
Multi$^3$Net was trained using the \textit{Adam} optimization algorithm \citep{kingma2014adam} with a learning rate of $10^{-2}$. 
The network parameters are optimized using a cross entropy loss $$H(\hat{\mathbf{y}},\mathbf{y}) = -\sum_{i} \mathbf{y}_i \log(\hat{\mathbf{y}}_i),$$ between ground truth $\mathbf{y}$ and predictions $\hat{\mathbf{y}}$. We anneal the learning rate according to the poly policy ($\text{power}=0.9$) introduced in \citet{chen2018deeplab} and stop training once the loss converges. For each batch, we randomly sample 8 tiles of size 960m$\times$960m  (corresponding to 96px$\times$96px optical and 192px$\times$192px radar images) from the dataset. We augment the training dataset by randomly rotating and flipping the image vertically and horizontally in order to create additional samples.
To segment flooded buildings with Multi$^3$Net, we first pre-train the network on building footprints. We then use the resulting weights for network initialization and train Multi$^3$Net on the footprints of flooded buildings.


\section{Data}

\begin{figure*}
	\tikzsetnextfilename{groundtruthexamples}
    \begin{subfigure}{.33\textwidth}
        \includegraphics[width=\textwidth]{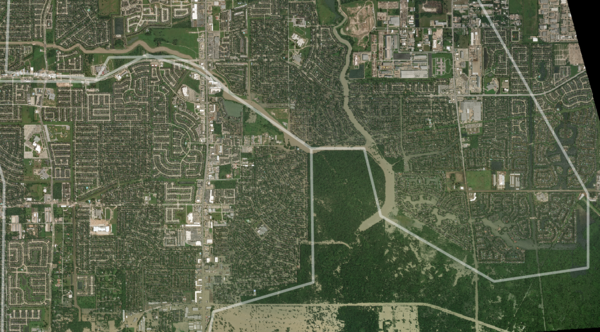}
        \caption{VHR image with partition boundaries.}
        \label{fig:aoi}
    \end{subfigure}
    \begin{subfigure}{.33\textwidth}
        \includegraphics[width=\textwidth]{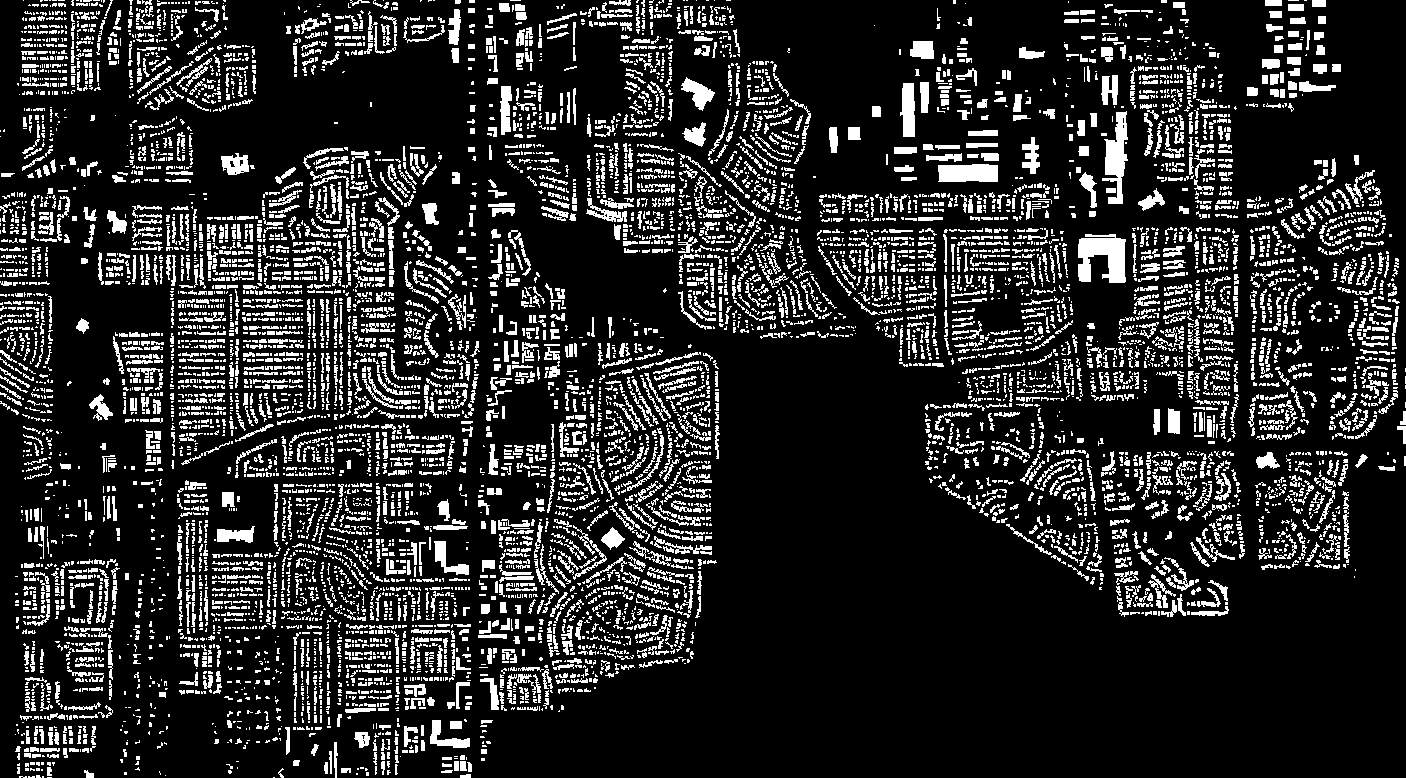}
        \caption{OpenStreetMap building footprints.}
        \label{fig:buildings}
    \end{subfigure}
    \begin{subfigure}{.33\textwidth}
        \includegraphics[width=\textwidth]{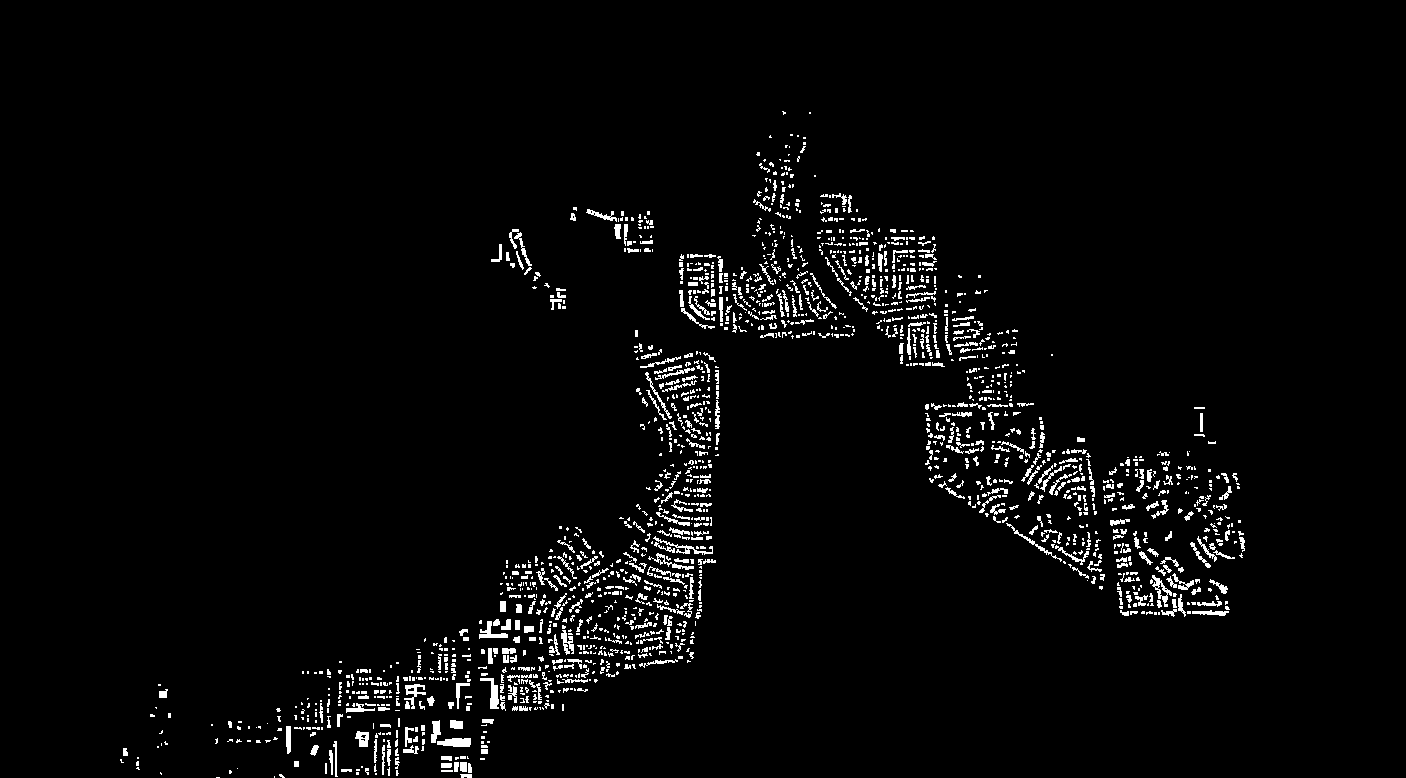}
        \caption{Annotated flooded buildings.}
        \label{fig:floodedbuildings}
    \end{subfigure}
    \caption{%
        Images illustrating (a) the size and extent of the dataset, (b) available rasterized ground truth annotations as OpenStreetMap building footprints, and (c) expert-annotated labels of flooded buildings c).
    }
    \label{fig:collectionaois}
\end{figure*}


\subsection{Area of Interest}

We chose two neighboring, non-overlapping districts of Houston, Texas as training and test areas. 
Houston was flooded in the wake of Hurricane Harvey, a category 4 hurricane that formed over the Atlantic on August 17, 2017, and made landfall along the coast of the state of Texas on August 25, 2017. The hurricane dissipated on September 2, 2017.
In the early hours of August 28, extreme rainfalls caused an `uncontrolled overflow’ of Houston's Addicks Reservoir and flooded the neighborhoods of `Bear Creek Village’, `Charlestown Colony’, `Concord Bridge’, and `Twin Lakes’.


\subsection{Ground Truth}

We chose this area of interest because accurate building footprints for the affected areas are publicly available through OpenStreetMap.
Flooded buildings have been manually labeled through crowdsourcing as part of the DigitalGlobe Open Data Program \citep{DigitalGlobeOpenData}.
When preprocessing the data, we combine the building footprints obtained from OpenStreetMap with point-wise annotations from DigitalGlobe to produce the ground truth map shown in \Cref{fig:floodedbuildings}.
The geometry collections of buildings (shown in \Cref{fig:buildings}) and flooded buildings (shown in \Cref{fig:floodedbuildings}) are then rasterized to create 2m or 10m pixel grids, depending on the satellite imagery available.
\Cref{fig:aoi} shows a very high-resolution image of the area of interest overlaid with boundaries for the East and West partitions used for training and testing, respectively.


\subsection{Data Preprocessing}

In Section \secref{sec:satellitedata}, we described the properties of short-wavelength optical and long-wavelength radar imagery.
For Sentinel-2 optical data, we use \textit{top-of-atmosphere} reflectances without applying further atmospheric corrections to minimize the amount of optical preprocessing need for our approach.
For radar data, however, preprocessing of the raw data is necessary to obtain numerical values that can be used as network inputs.
A single radar `pixel' is expressed as a complex number $z$ and composed of a real in-phase, $\Re(z)$, and an imaginary quadrature component of the reflected electromagnetic signal, $\Im(z)$.
We use \textit{single look complex} data to derive the radar intensity and coherence features.
The intensity, defined as $I \equiv z^2 = \Re(z)^2 + \Im(z)^2$, contains information about the magnitude of the surface-reflected energy. The radar images are preprocessed according to \citet{ulaby2014}:
(1) We perform \textit{radiometric calibration} to compensate for the effects of the sensor's relative orientation to the illuminated scene and the distance between them.
(2) We reduce the noise induced by electromagnetic interference, known as \textit{speckle}, by applying a spatial averaging kernel, known as \textit{multi-looking} in radar nomenclature.
(3) We normalize the effects of the terrain elevation using a digital elevation model, a process known as \textit{terrain correction}, where a coordinate is assigned to each pixel through \textit{georeferencing}.
(4) We average the intensity of all radar images over an extended temporal period, known as \textit{temporal multi-looking}, to further reduce the effect of speckle on the image.
(5) We calculate the \textit{interferometric coherence} between images, $\mathbf{z}_t$, at times $t=1,2$,
\begin{equation}
    \label{eq:coherence}
    \gamma = \frac{\E[\mathbf{z}_1\mathbf{z}_2^\ast]}
            {\sqrt{\E[|\mathbf{z}_1|^2]\E[|\mathbf{z}_2|^2]}},
\end{equation}
where $\mathbf{z}_t^\ast$ is the complex conjugate of $\mathbf{z}_t$ and expectations are computed using a local \textit{boxcar-function}.
The coherence is a local similarity metric \citep{Zebker1992DecorrelationII} able to measure changes between pairs of radar images.


\subsection{Network Inputs}

We use medium-resolution satellite imagery with a ground resolution of 5m--10m, acquired before and after disaster events, along with very high-resolution post-event images with a ground resolution of 0.5m.
Medium-resolution satellite imagery is publicly available for any location globally and acquired weekly by the European Space Agency.

For radar data, we construct a three-band image consisting of the intensity, multitemporal filtered intensity, and interferometric coherence.
We compute the intensity of two radar images obtained from Sentinel-1 sensors in stripmap mode with a ground resolution of 5m for August 23 and September 4, 2017.
Additionally, we calculate the interferometric coherence for an image pair without flood-related changes acquired on June 6 and August 23, 2017, as well as for an image pair with flood-induced scene changes acquired on August 23 and September 4, 2017, using \Cref{eq:coherence}.
Examples of coherence images generated this way are shown in \Cref{fig:tile:s1pre,fig:tile:s1post}.
As the third band of the radar input, we compute the multitemporal intensity by averaging all Sentinel-1 radar images from 2016 and 2017. This way, speckle noise affecting the radar image can be reduced.
We merge the intensity, multitemporal filtered intensity, and coherence images obtained both pre- and post-disaster into separate three-band images.
The multi-band images are then fed into the respective network streams.

\Cref{fig:tile:s2pre,fig:tile:s2post} show pre- and post-event images obtained from the Sentinel-2 satellite constellation on August 20 and September 4, 2017.
Sentinel-2 measures the surface reflectances in 13 spectral bands with 10m, 20m, and 60m ground resolutions.
We apply bilinear interpolations to the 20m band images to obtain an image representation with 10m ground resolution.
To obtain finer image details, such as building delineations, we use very high-resolution post-event images obtained through the DigitalGlobe Open Data Program (see \Cref{fig:tile:vhr}).
The very high-resolution image used in this work was acquired on August 31, 2017, and contains three spectral bands (red, green, and blue), each with a 0.5m ground resolution.

Finally, we extract rectangular tiles of size 960m$\times$960m from the set of satellite images to use as input samples for the network.
This tile extraction process is repeated every 100m in the four cardinal directions to produce overlapping tiles for training and testing, respectively. The large tile overlap can be interpreted as an offline data augmentation step.

\def\imagewidth{2.8cm}
\begin{figure*}[h]
	\tikzsetnextfilename{qualitativeresults}
    \begin{tikzpicture}[node distance=.5em, text depth=0pt]
    
        \node[inner sep=0, label=Sentinel-2 Input] (a)
        	{\includegraphics[width=\imagewidth]{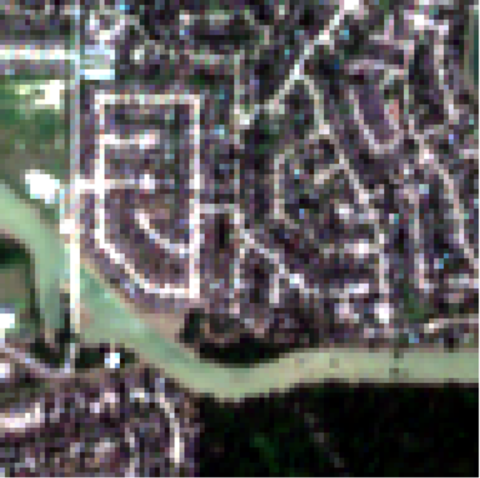}};
        \node[inner sep=0, right=of a, label=Target (10m)] (b) {\includegraphics[width=\imagewidth]{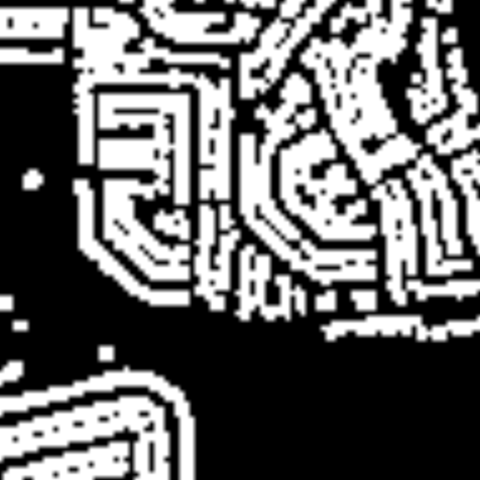}};
        \node[inner sep=0, right=of b, label=Prediction] (c) {\includegraphics[width=\imagewidth]{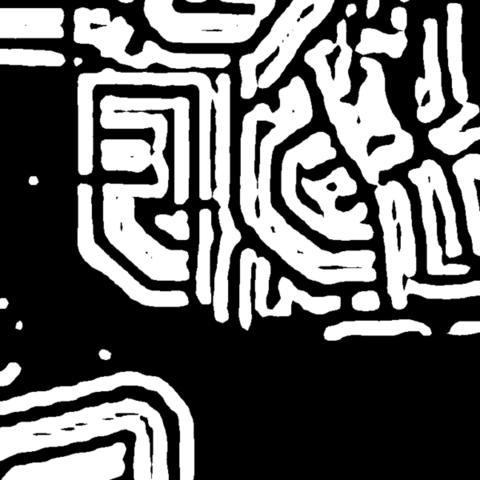}};
        \node[inner sep=0, right= of c, label=VHR Input] (a2) {\includegraphics[width=\imagewidth]{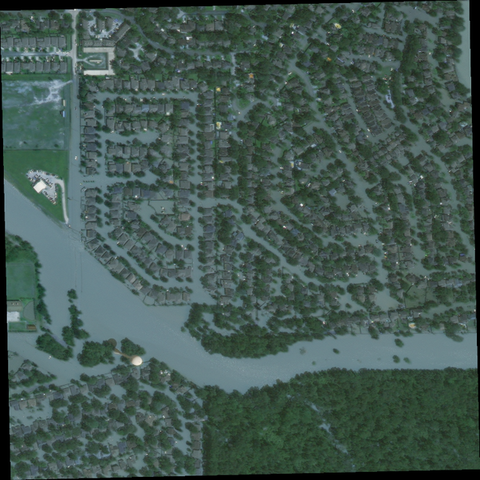}};
        \node[inner sep=0, right=of a2, label=Target (2m)] (b2) {\includegraphics[width=\imagewidth]{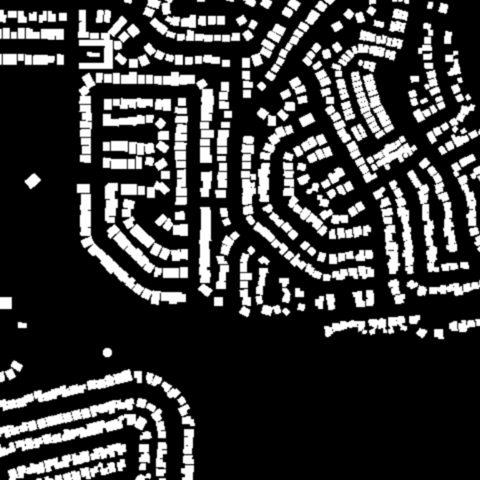}};
        \node[inner sep=0, right=of b2, label=Prediction] (c2) {\includegraphics[width=\imagewidth]{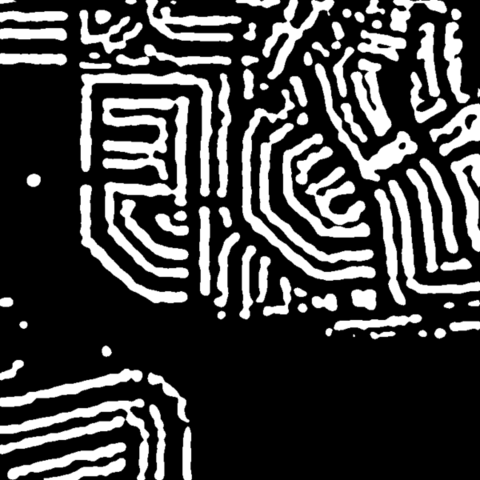}};
        
        \coordinate (break) at ($(c)!0.5!(a2)$);
        \draw[thick] ($ (break) - (0,1.6cm) $) -- ($ (break) + (0,1.6cm) $) node[at end, above=3em]{};
        
    \end{tikzpicture}
    \caption{Prediction targets and prediction results for building footprint segmentation using Sentinel-1 and Sentinel-2 inputs fused at a 10m resolution (left panel) and using Sentinel-1, Sentinel-2, and VHR inputs fused at a 2m resolution (right panel).}
    \label{fig:OneModBuildingS2VHRQual}
\end{figure*}


\section{Experiments \& Results}
In this section, we present quantitative and qualitative results for the segmentation of building footprints and flooded buildings. We show that fusion-based approaches consistently outperform models that only incorporate data from single sensors.


\subsection{Evaluation Metrics}

We segment building footprints and flooded buildings and compare the results to state-of-the-art benchmarks. To assess model performance, we report the \textit{Intersection over Union} (IoU) metric, which is defined as the number of overlapping pixels labeled as belonging to a certain class in both target image and prediction divided by the union of pixels representing the same class in target image and prediction. We use it to assess the predictions of building footprints and flooded buildings obtained from the model. We report this metric using the acronym `bIoU'. Represented as a confusion matrix, $\text{bIoU}\equiv \text{TP}/(\text{FP}+\text{TP}+\text{FN})$, where TP $\equiv$ True Positives, FP $\equiv$ False Positives, TN $\equiv$ True Negatives, and FN $\equiv$ False Negatives. Conversely, the IoU for the background class, in our case denoting `not a flooded building', is given by $\text{TN}/(\text{TN}+\text{FP}+\text{FN})$. Additionally, we report the mean of (flooded) building and background IoU values, abbreviated as `mIoU'.
We also compute the pixel accuracy $A$, the percentage of correctly classified pixels, as $A \equiv (\text{TP}+\text{TN}) / (\text{TP}+\text{FP}+\text{TN}+\text{FN})$.


\subsection{Building Footprint Segmentation: Single Sensors}
\label{sec:exp:inria}

We tested our model on the auxiliary task of building footprint segmentation. The wide applicability of this task has led to the creation of several benchmark datasets, such as the DeepGlobe \citep{demir2018DeepGlobe}, SpaceNet \citep{van2018spacenet}, and INRIA aerial labels datasets \citep{maggiori2017dataset}, all containing very high-resolution RGB satellite imagery.
\Cref{tab:inria} shows the performance of our model on the Austin partition of the INRIA aerial labels dataset.
\Citet{maggiori2017convolutional} use a fully convolutional network \citep{long2015fully} to extract features that were concatenated and classified by a second multilayer perceptron stream.
\Citet{ohleyer2018footprint} employ a Mask-RCNN \citep{he2017mask} instance segmentation network for building footprint segmentation.

Using only very high-resolution imagery, Multi$^3$Net performed better than current state-of-the-art models, reaching a bIoU 7.8\% higher than \citet{ohleyer2018footprint}. Comparing the performance of our model for different single-sensor inputs, we found that predictions based on very high-resolution images achieved the highest building IoU score, followed by predictions based on Sentinel-2 medium-resolution optical images, suggesting that optical bands contain more relevant information for this prediction task than radar images.

\begin{table}[h]
\begin{tabularx}{\linewidth}{X c c} 
 \toprule
 \textbf{Model} & \textbf{bIoU} & \textbf{Accuracy} 
 \\
 \cmidrule(lr){1-1}\cmidrule(lr){2-2}\cmidrule(lr){3-3}
 \citet{maggiori2017convolutional} & 61.2\% & 94.2\% \\
 \citet{ohleyer2018footprint} & 65.6\% & 94.1\% \\
\textbf{Multi$^\textbf{3}$Net} & \bfseries73.4\% & \bfseries95.7\% \\
 \bottomrule
\end{tabularx}
\caption{Building footprint segmentation results based on VHR images of the Austin partition of the INRIA aerial labels dataset \citep{maggiori2017dataset}.}
\label{tab:inria}
\end{table}

\def\imagewidth{2.8cm}
\begin{figure*}
	\tikzsetnextfilename{qualitativeexamples2}
\centering
    \begin{tikzpicture}[node distance=.5em, text depth=0pt]
        \node[inner sep=0, label=VHR Input] (a) {\includegraphics[width=0.19\textwidth]{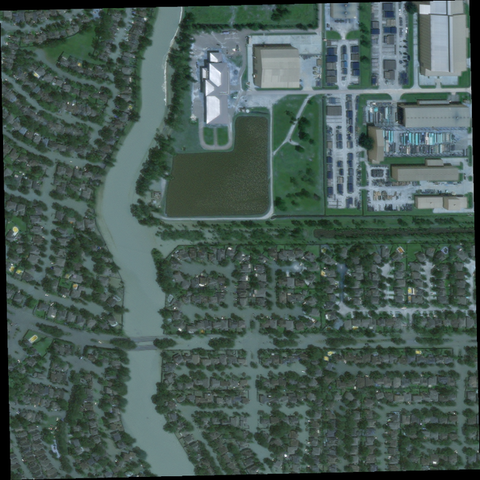}};
        \node[inner sep=0, right=of a, label=Target] (b) {\includegraphics[width=0.19\textwidth]{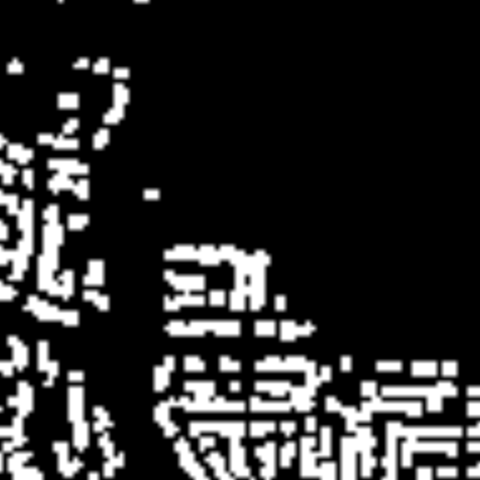}};
        \node[inner sep=0, right=of b, label= Fusion Prediction ] (c) {\includegraphics[width=0.19\textwidth]{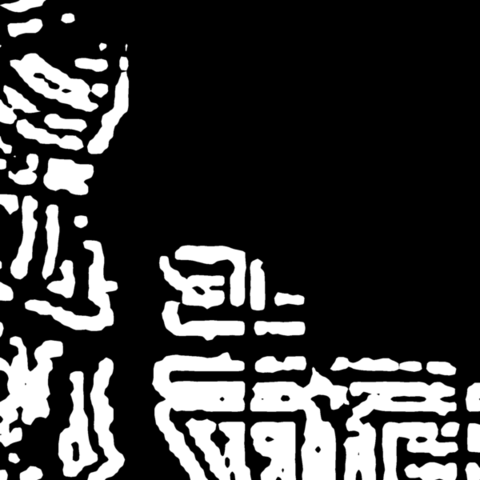}};
        \node[inner sep=0, right=of c, label=VHR-only Prediction] (d) {\includegraphics[width=0.19\textwidth]{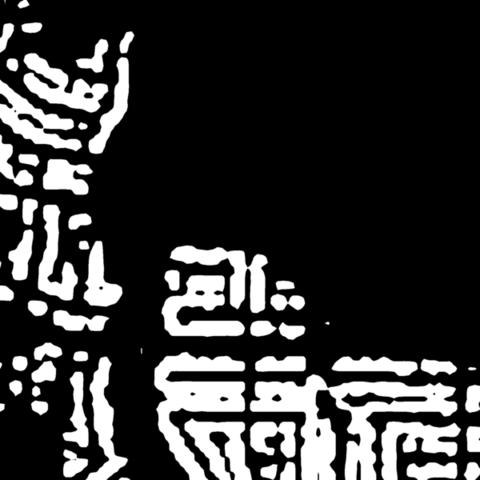}};
        \node[inner sep=0, right=of d, label=Overlay] (e) {\includegraphics[width=0.19\textwidth]{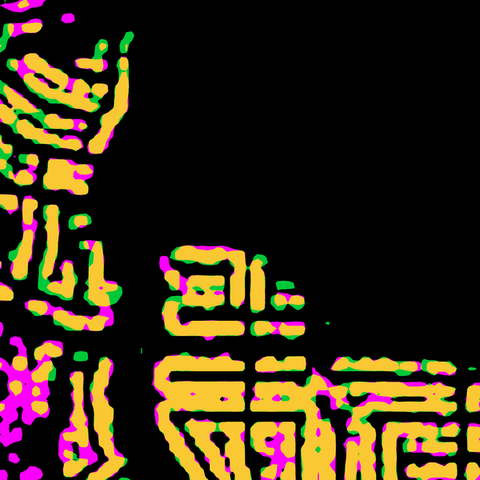}};
    \end{tikzpicture}
    \caption{Comparison of predictions for the segmentation of flooded buildings for fusion-based and VHR-only models. In the overlay image, predictions added by the fusion are marked in magenta, predictions that were removed by the fusion are marked in green, and predictions present in both are marked in yellow.}
    \label{fig:FuseModDamageQual}
\end{figure*}


\subsection{Building Footprint Segmentation: Image Fusion}
\label{sec:buildingfootprint:fusionmodalities}

Fusing multiresolution and multisensor satellite imagery further improved the predictive performance. The results presented in \Cref{tab:FuseModBuilding} show that the highest accuracy was achieved when all data sources were fused. We also compared the performance of Multi$^3$Net to the performance of a baseline U-Net data fusion architecture, which has been successful at recent satellite imagery segmentation competitions, and found that Multi$^3$Net outperformed the U-Net baseline on building footprint segmentation for all input types (see \textit{Appendix} for details).

\Cref{fig:OneModBuildingS2VHRQual} shows qualitative building footprint segmentation results when fusing images from multiple sensors. Fusing Sentinel-1 and Sentinel-2 data produced highly accurate predictions (76.1\% mIoU), only surpassed by predictions obtained by fusing Sentinel-1, Sentinel-2, and very high-resolution imagery (79.9\%).

\begin{table}[h]
\begin{tabularx}{\linewidth}{X c c c} 
 \toprule
 \textbf{Data} & \textbf{mIoU} & \textbf{bIoU} & \textbf{Accuracy} \\ 
 \cmidrule(lr){1-1}\cmidrule(lr){2-2}\cmidrule(lr){3-3}\cmidrule(lr){4-4}
  S-1 & 69.3\% & 63.7\% & 82.6\% \\
 S-2 & 73.1\% & 66.7\% & 85.4\% \\
 VHR & 78.9\% & 74.3\% & 88.8\% \\
 S-1 + S-2 & 76.1\% &  70.5\% & 87.3\% \\
 S-1 + S-2 + VHR & \bfseries 79.9\% & \bfseries 75.2\% & \bfseries89.5\% \\
\bottomrule
\end{tabularx}
\caption{Results for the segmentation of building footprints using different input data in Multi$^3$Net.}
\label{tab:FuseModBuilding}
\end{table}


\subsection{Segmentation of Flooded Buildings with Multi$^\mathbf{3}$Net}
\label{sec:quant:flooddetection}

To perform highly accurate segmentation of flooded buildings, we add multitemporal input data obtained from Sentinel-1 and Sentinel-2 to our fusion network. \Cref{tab:FuseModDamage} shows that using multiresolution and multisensor data across time yielded the best performance (75.3\% mIoU) compared to other model inputs. Furthermore, we found that, despite the significant difference in resolution between medium- and very high-resolution imagery, fusing globally available medium-resolution images from Sentinel-1 and Sentinel-2 also performed well, reaching a mean IoU score of 59.7\%. These results highlight one of the defining features of our method: A segmentation map can be produced as soon as at least a single satellite acquisition has been successful and subsequently be improved upon once additional imagery becomes available, making our method flexible and useful in practice (see \Cref{tab:FuseModBuilding}). We also compared Multi$^3$Net to a U-Net fusion model and found that Multi$^3$Net performed significantly better, reaching a building IoU score of 75.3\% compared to a bIoU score of only 44.2\% for the U-Net baseline.

\Cref{fig:FuseModDamageQual} shows predictions for the segmentation of flooded buildings obtained from the very high-resolution-only and full-fusion models. The overlay image shows the differences between the two predictions.
Fusing images from multiple resolutions and multiple sensors across time eliminates the majority of false positives and helps delineate the shape of detected structures more accurately.
The flooded buildings in the bottom left corner, highlighted in magenta, for example, were only detected using multisensor input.

\begin{table}[h]
\begin{tabularx}{\linewidth}{X c c c} 
 \toprule
 \textbf{Data} & \textbf{mIoU} & \textbf{bIoU} & \textbf{Accuracy} \\
 \cmidrule(lr){1-1}\cmidrule(lr){2-2}\cmidrule(lr){3-3}\cmidrule(lr){4-4}
 S-1 & 50.2\% & 17.1\% & 80.6\% \\
 S-2 & 52.6\% & 12.7\% & 81.2\% \\
 VHR & 74.2\% & 56.0\% & 93.1\% \\
 S-1 + S-2 & 59.7\% & 34.1\% & 86.4\% \\
 S-1 + S-2 + VHR & \bfseries75.3\% & \bfseries57.5\% & \bfseries93.7\% \\
\bottomrule
\end{tabularx}
\caption{Results for the segmentation of flooded buildings using different input data in Multi$^3$Net.}
\label{tab:FuseModDamage}
\end{table}


\section{Conclusion}

In disaster response, fast information extraction is crucial for first responders to coordinate disaster relief efforts, and satellite imagery can be a valuable asset for rapid mapping of affected areas.
In this work, we introduced a novel end-to-end trainable convolutional neural network architecture for fusion of multiresolution, multisensor optical and radar satellite images that outperforms state-of-the-art models for segmentation of building footprints and flooded buildings.

We used state-of-the-art pyramid sampling pooling \citep{zhao2017pyramid} to aggregate spatial context and found that this architecture outperformed fully convolutional networks \citep{maggiori2017convolutional} and Mask-RCNNs \citep{ohleyer2018footprint} on building footprint segmentation from very high-resolution images.
We showed that building footprint predictions obtained by only using publicly-available medium-resolution radar and optical satellite images in Multi$^3$Net almost performs on par with building footprint segmentation models that use very high-resolution satellite imagery \citep{Bischke2017MultiTaskLF}.
Building on this result, we used Multi$^3$Net to segment flooded buildings, fusing multiresolution, multisensor, and multitemporal satellite imagery, and showed that full-fusion outperformed alternative fusion approaches. This result demonstrates the utility of data fusion for image segmentation and showcases the effectiveness of Multi$^3$Net's fusion architecture. Additionally, we demonstrated that using publicly available medium-resolution Sentinel imagery in Multi$^3$Net produces highly accurate flood maps.

Our method is applicable to different types of flood events, easy to deploy, and substantially reduces the amount of time needed to produce highly-accurate flood maps.
We also release the first open-source dataset of fully preprocessed and labeled multiresolution, multispectral, and multitemporal satellite images of disaster sites along with our source code, which we hope will encourage future research into image fusion for disaster relief.


\section{Acknowledgements}

This research was conducted at the Frontier Development Lab (FDL), Europe. The authors gratefully acknowledge support from the European Space Agency, NVIDIA Corporation,  Satellite Applications Catapult, and Kellogg College, University of Oxford.


\newpage

\bibliography{references}
\bibliographystyle{include/aaai}

\appendix

\onecolumn

\section{\LARGE Appendix}
\vspace{0.5cm}
\subsection{A1. Training \& Model Evaluation Details}

\begin{figure*}[h]
\begin{center}
    \includegraphics[width=0.8\textwidth]{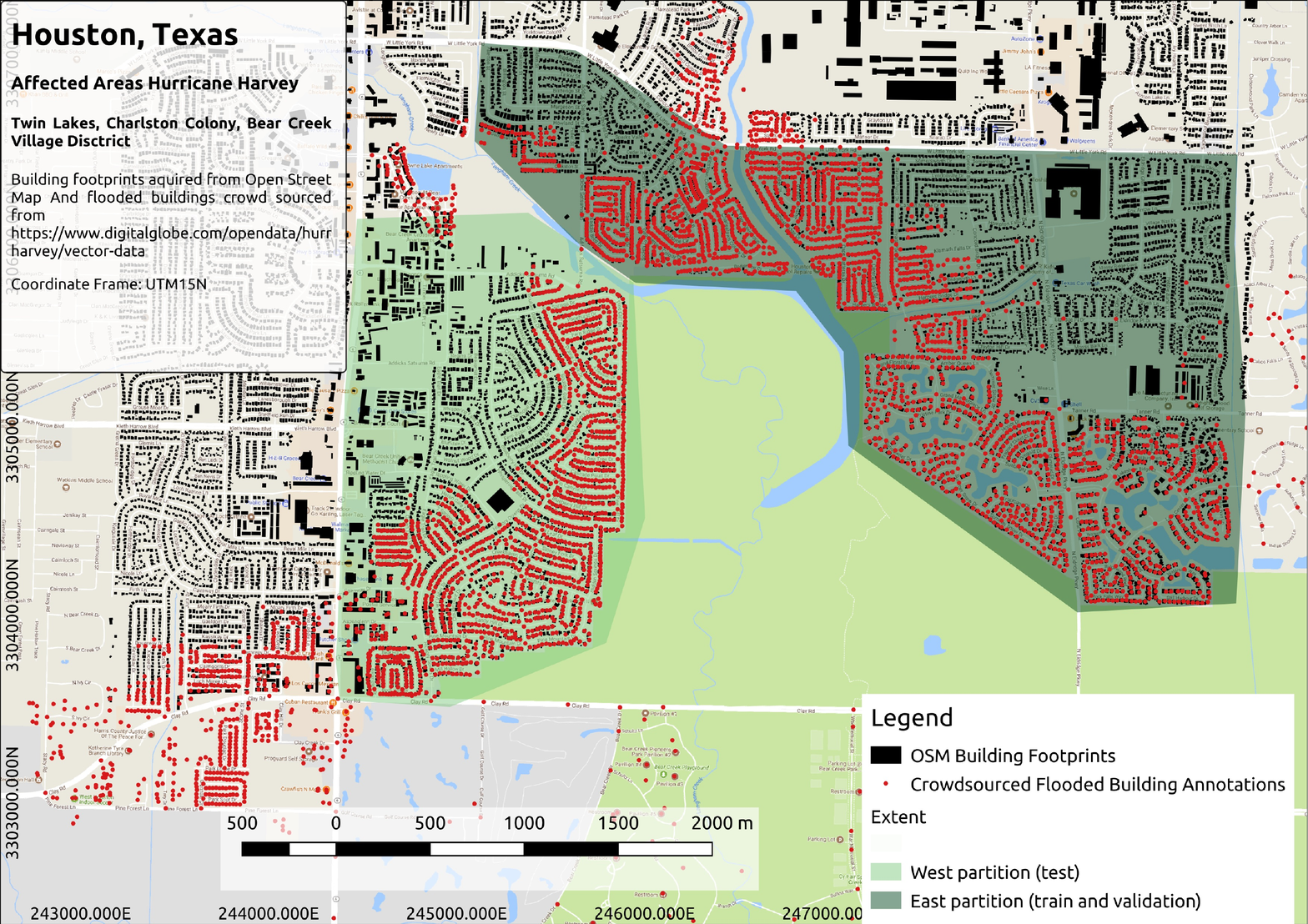}
    \vspace{-.5em}
    \caption{Detailed map of the area of interest. The shaded regions are the East and West partitions used for training and testing the model, respectively. Flooded buildings are highlighted in red.}
    \label{fig:partition_map}
\end{center}
\end{figure*}

To train our models, we divided the area of interest into two partitions (i.e. non-overlapping subsets) covering two different neighborhoods, as shown in \Cref{fig:aoi} and \Cref{fig:partition_map}. We randomly divided the East partition into a training and a validation set at a 4:1 split. The model  hyperparameters were optimized on the validation set. All model evaluations presented in this work were performed on the spatially separate test dataset.


\subsection{A2. Additional Experiments}
\label{app:exp}

We compared the performance of Multi$^3$Net to the performance of a baseline U-Net data fusion architecture, which has been successful at recent satellite image segmentation competitions, and found that our model outperformed the U-Net baseline on building footprint segmentation for all input types (see \Cref{tab:OneModBuilding}). We also compared the performance between Multi$^3$Net and a baseline U-Net fusion architecture on the segmentation of flooded buildings and found that our method performed significantly better, reaching a building IoU (bIoU) score of 75.3\% compared to a bIoU score of 44.2\% for the U-Net baseline.

\begin{table}[h]
\vspace*{-0.2cm}
\begin{tabularx}{\linewidth}{c X c c c} 
 \toprule
  \textbf{Model} & \textbf{Data} & \textbf{mIoU} & \textbf{bIoU} & \textbf{Accuracy} \\
 \cmidrule(lr){1-1}
 \cmidrule(lr){2-2}
 \cmidrule(lr){3-3}
 \cmidrule(lr){4-4}
 \cmidrule(lr){5-5}

 \textbf{Multi$^\textbf{3}$Net} & Sentinel-1 + Sentinel-2 & 76.1\% &  70.5\% & 87.3\% \\
 & VHR & 78.9\% & 74.3\% & 88.8\% \\
 & Sentinel-1 + Sentinel-2 + VHR & \bfseries 79.9\% & \bfseries 75.2\% & \bfseries89.5\% \\
\midrule
 \textbf{U-Net} & Sentinel-1 + Sentinel-2 & - & 60\% & 88\% \\
 & VHR & - & 38\% & 77\% \\
 & Sentinel-1 + Sentinel-2 + VHR & - & \bfseries 73\% & \bfseries  89\% \\
  \bottomrule
\end{tabularx}
\caption{Building footprint segmentation results for Multi$^3$Net and a U-Net baseline.}
\vspace{-0.3cm}
\label{tab:OneModBuilding}
\end{table}

\end{document}